\pdfoutput=1

\documentclass[11pt]{article}

\usepackage{acl}

\usepackage{times}
\usepackage{latexsym}
\usepackage{graphicx}
\usepackage{amsfonts} 
\usepackage{amsmath}
\usepackage{amssymb}
\usepackage{booktabs}

\usepackage[T1]{fontenc}

\usepackage[utf8]{inputenc}
\usepackage[activate={true,nocompatibility},final,tracking=false,kerning=true,spacing=true,factor=1100,stretch=20,shrink=30]{microtype}

\usepackage{inconsolata}

\usepackage{enumitem}
\setlist[itemize]{leftmargin=*}

\usepackage{fontawesome}
\usepackage{xcolor} 
\usepackage[bordercolor=gray!0,backgroundcolor=red!10,linecolor=red!50,textsize=tiny,textwidth=.9in,obeyFinal]{todonotes}

\newcommand{\studentNST}{\texttt{Student\textsubscript{NST}}}
\newcommand{\studentCRD}{\texttt{Student\textsubscript{CRD}}}

\newcommand{\BERT}{\texttt{BERT}}
\newcommand{\wikiBERT}{\texttt{wikiBERT}}
\newcommand{\spokenBERT}{\texttt{transcription-BERT}}

\newcommand{\lossNST}{\mathcal{L}_{\texttt{KD}}^{\texttt{NST}}}
\newcommand{\lossMMD}{\texttt{MMD}}
\newcommand{\lossCRD}{\mathcal{L}_{\texttt{KD}}^{\texttt{CRD}}}

\title{\textit{Teach me with a Whisper}: Enhancing Large Language Models  \\ 
for Analyzing Spoken Transcripts using Speech Embeddings}

\author{Fatema Hasan$^1$, Yulong Li$^2$, James Foulds$^1$,  Shimei Pan$^1$, Bishwaranjan Bhattacharjee$^2$ \AND
$^1$Department of Information Systems \\
University of Maryland Baltimore County \\
Maryland 21250, USA \And 
$^2$IBM Research AI \\
Yorktown Heights\\
NY 10598, USA \AND
\texttt{fhasan1,jrfoulds,shimei@umbc.edu} \And 
\texttt{yulongl,bhatta@us.ibm.com}
}

\begin{document}
\maketitle
\begin{abstract}
Speech data has rich acoustic and para-linguistic information with important cues for understanding a  speaker's tone, emotion, and intent, 
yet traditional large language models such as BERT do not incorporate this information.
There has been an increased interest in multi-modal language models leveraging audio and/or visual information and text. However, current multi-modal language models require both text and audio/visual data streams during inference/test time.  
In this work, we propose a methodology for training language models leveraging spoken language audio data but without requiring the audio stream during prediction time. 
This leads to an improved language model for analyzing spoken transcripts while avoiding an audio processing overhead at test time. 
We achieve this via an audio-language knowledge distillation framework,  where we transfer acoustic and paralinguistic information from a pre-trained speech embedding (OpenAI Whisper) teacher model to help train a student language model on an audio-text dataset. In our experiments, the student model achieves consistent improvement over traditional language models on tasks analyzing spoken transcripts. 

\end{abstract}

\section{Introduction}
Large language models (LLMs) such as BERT~\citep{devlin2018bert} have achieved remarkable success in various natural language processing (NLP) tasks by training on vast text corpora (e.g. Wikipedia), enabling them to capture and model complex linguistic patterns. 
However, these models typically do not incorporate an important aspect of human communication: the rich acoustic and para-linguistic information conveyed through speech, encompassing a wide range of nonverbal cues and contextual nuances~\citep{gibson1993tools} that cannot be captured 
through written text \citep{kumar2021bert}. These cues include variations in tone, pitch, intonation, emphasis, pauses, and other vocal characteristics that add layers of meaning and convey emotional states, speaker attitudes, and intentions~\citep{manning2014stanford}. 
By solely relying on written text, traditional language models miss out on these crucial signals.

In contrast, incorporating and leveraging rich audio and para-linguistic information within models can greatly enhance their ability to comprehend and generate language, capturing the holistic nature of human communication. 
Access to such 
information can enable language models to interpret sarcasm~\citep{castro2019towards}, detect emotions and sentiments~\citep{zadeh2018multimodal, busso2008iemocap}, understand the speaker's emphasis or urgency~\citep{tsai2019multimodal}, and identify other nonverbal communication aspects that significantly impact the overall meaning and interpretation of spoken language. 
Incorporating these data into language models has the potential to improve their understanding of contextually appropriate and authentic responses in informal settings. 

Recognizing the inherent value of integrating different modalities, there has been a growing interest in leveraging the combined power of audio and/or visual information with textual information to create multi-modal representations~\citep{yang2022code, tsai2019multimodal} in recent years.
However, due to the inherent design, during the actual usage (i.e. inference after deployment) the model requires the user to provide both the text and the audio/visual data stream.
This reliance on multiple modalities during inference after deployment presents challenges in certain scenarios. For instance, in resource-constrained environments or applications where only textual input is available, the full benefits of these multi-modal models may not be realized. Additionally, during the actual deployment or usage of these models, they need to process and integrate data from multiple sources simultaneously resulting in increased computational cost and inefficient processing pipelines.

\begin{figure}[t]
\centering
\includegraphics[width=0.5\textwidth]{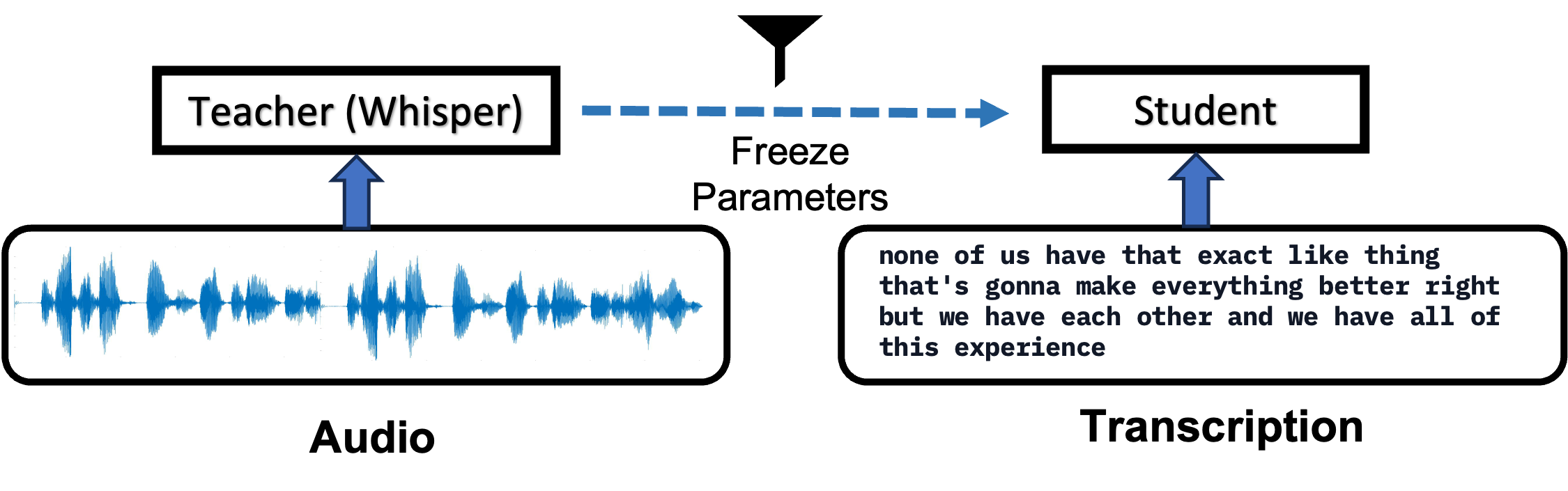}
\caption{Overview of our system.}
\label{fig:overview}
\end{figure}

Recent work~\citep{tan2020vokenization} has sought to capitalize on the availability of visually-grounded language datasets (e.g. MS Coco~\citep{lin2014microsoft}) with the aim of improving the representation of language. However, it is important to note that the existence of comparable datasets specifically designed for the audio-text modality is severely constrained or non-existent. To overcome these limitations, we propose a language model trained using human-spoken audio-text data but without requiring audio at inference time. To achieve this, we present an audio-language knowledge distillation method to train a text-only language model. 
We leverage a pre-trained frozen audio embedding model (OpenAI whisper) as the teacher model to transfer its knowledge to train a student model on an audio-text dataset. 
In our knowledge-distillation system (see Figure~\ref{fig:overview}), the use of large-scale human-spoken spontaneous audio and its transcribed text dataset helps us learn diverse and richer vocabularies of natural and informal human-spoken language. To summarize, the main contributions of our paper are:
\begin{itemize}
    \item 
    We develop a method to train a text-only LLM while leveraging acoustic and paralinguistic information.
    \item 

    We design an audio-language knowledge distillation approach that leverages pre-trained audio embeddings as a teacher model and transfers its knowledge to train a text-based student model. 
    \item We empirically show the impact of incorporating audio and paralinguistic features to text on sentiment analysis and emotion recognition tasks. 
\end{itemize}




\section{Related Works}

\subsection{Knowledge Distillation}
Knowledge distillation (KD) \cite{hinton2015distilling} is a class of methods for training a ``student'' model 
based on 
a teacher model where the KD aims to match the student's predictions to the teacher's (usually pretrained) predictions. It has been successfully demonstrated in various tasks such as machine translation \citep{DBLP:journals/corr/KimR16a}, visual recognition \citep{He_2019_CVPR}, and speech recognition \citep{46282} with applications of interpretability and model compression.

Recent development of KD has expended its applications from single modality to cross modalities. For example, \citet{tang2021vidlankd} combine contrastive learning and KD to transfer the knowledge from video to text. They train a multimodal teacher model on a video-text dataset and then distill its knowledge to a student language model with a text dataset. \citet{kim2022cross} apply cross-attention between audio and text features and cross-modal distillation to improve multi-class emotion classification tasks. In our work, we adopt a similar strategy to \citet{tang2021vidlankd} but transfer the acoustic and paralinguistic information from audio embedding to text embedding during pretraining. Instead of training our own teacher model, we utilize pretrained audio embedding model to improve training efficiency. Different from \citep{kim2022cross} which requires both audio and text data, our model works with only text data during inference. 

\subsection{Multimodal Learning}
Inspired by the success of language pretraining with transformer models, training with multimodal data has attracted more and more attention. Not only it can improve multi-modal downstream tasks \citep{antol2015vqa, 7780940} when using image-text \citep{DBLP:journals/corr/abs-1909-11740, DBLP:journals/corr/abs-1908-06066} or video-text data \cite{DBLP:journals/corr/abs-2005-00200, DBLP:journals/corr/abs-1912-06430}, multimodal learning can also improve traditional single-modal tasks such as language understanding \cite{tang2021vidlankd}. \citet{tan2020vokenization} propose a visually-supervised language model which predicts a visualized token for each input text token during pretraining, therefore providing grounding to the language model.

\begin{figure}[!tbh]
\centering
\includegraphics[width=0.48\textwidth]{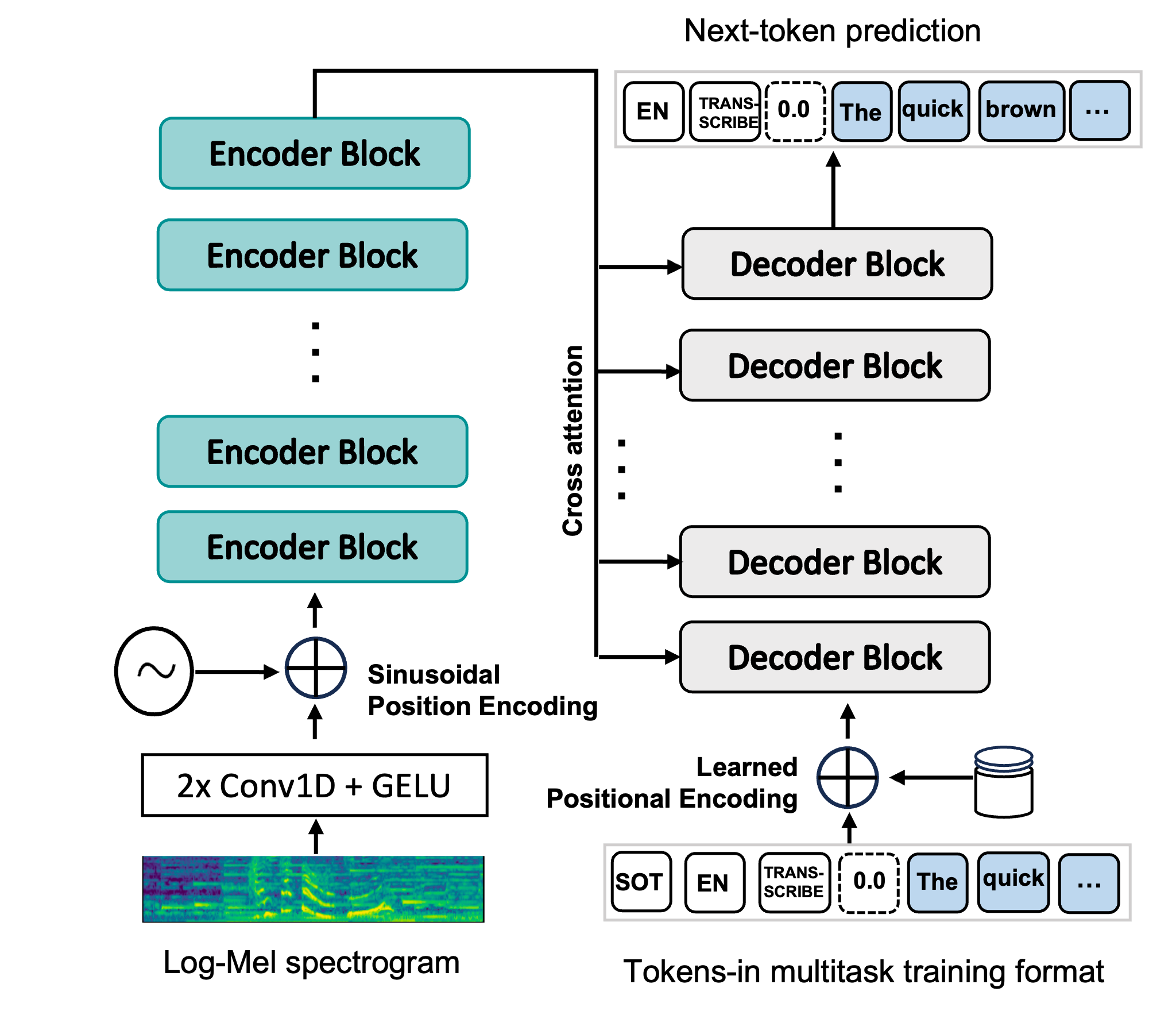}
\caption{Overview of the OpenAI-Whisper model. 
}
\label{fig:whisper}
\end{figure}

Audio data has also been used in multimodal learning \citep{kang2022self, chudasama2022m2fnet, siriwardhana2020multimodal}, \citet{DBLP:journals/corr/abs-1910-11559} propose a pretraining audio-and-text model for spoken question answering. \citet{DBLP:journals/corr/abs-1906-00295} introduce the Multimodal Transformer which learns representations directly from unaligned multimodal streams including vision, language and audio for human multimodal affection recognition task. However, most of the work requires multimodal data during inference, which is not always easy to acquire. \citet{yang2022code} develop a composable multimodal learning framework which works with both multimodal and single-modal data. Despite their more complicated design, the performance on  written formal language tasks (GLUE) was only marginally improved. Different from \citet{yang2022code}, our work aims to use audio information to enhance the analysis of spoken transcripts. 


\section{Audio-Language Knowledge Distillation}

\subsection{Method Overview}
We aim to learn a better language representation with the knowledge distilled from speech information. For this, we leverage an aligned multi-modal dataset $\mathcal{D}_{\texttt{{Audio-Text}}}:\{(a, x)\}$ (e.g., People’s Speech~\citep{galvez2021people}) where each audio $a$ sample is paired with its transcribed text $x$. 
We train our network with two objectives: (i) cross-modal knowledge distillation objective to transfer knowledge from the teacher to the student model and (ii) masked language modeling for the student model to induce its language understanding capabilities. Therefore, the overall loss function 
is:
\newcommand{\lossKD}{\mathcal{L}_\texttt{KD}}
\newcommand{\lossMLM}{\mathcal{L}_\texttt{MLM}}

\begin{equation}
\label{eqn:total_loss}
\mathcal{L} = \gamma \lossKD + \lossMLM 
\end{equation}
Here, $\gamma$ parameter controls the relative strength of $\lossKD$ with respect to $\lossMLM$.
Our teacher model is a pretrained speech embedding model (OpenAI Whisper) which we keep frozen during training. Our student model is a traditional transformer-based encoder model which we train from scratch.

\subsection{Teacher Model}
\label{sec:teacher}
In contrast to other multi-modal knowledge distillation approaches~\citep{tang2021vidlankd}, we refrain from training our own teacher model and instead leverage an existing, readily available pre-trained speech embeddings model known as Whisper~\citep{radford2022robust}. 
This strategic decision not only allows us to significantly reduce computational time and resource requirements (thereby improving efficiency), but it also ensures that we have a 
rich and diverse representation of the input audio samples which is crucial for a proficient teacher model. It is also worth noting that \textit{Whisper} outperforms~\cite{chemudupati2023transferability} all the available pre-trained speech embeddings models such as HuBERT~\cite{hsu2021hubert}, wav2vec2.0~\cite{baevski2020wav2vec}, and wavLM~\cite{chen2022wavlm} on the benchmark datasets.

\textit{Whisper} is an automatic speech recognition pre-trained model developed by OpenAI. It has undergone pretraining using 
680,000 hours of multilingual and multitask supervised data collected from the web. The architecture of Whisper is based on an encoder-decoder transformer. 
The encoder and decoder components have the same widths and numbers of transformer blocks. 
At first, the raw audio inputs are converted to a log-Mel spectrogram. 
This input representation is then fed through two convolution layers with GELU activation functions. 
To incorporate positional information, sinusoidal position embeddings are added to the output of the convolutional layers. 
The output is subsequently fed into the encoder blocks, resulting in a sequence of encoder hidden states.
Finally, the decoder predicts text tokens by learning positional embeddings and tying input-output token representations, considering both the previous tokens and the encoder hidden states. 
In our 
experiment, we only use the encoder blocks of Whisper as our teacher model. We utilize \texttt{whisper-small.en}, which comprises 12 layers of transformers containing 244 million parameters and has been trained solely on English speech-text data. 

For each audio sample $a$, we extract log-Mel spectrogram features $e^a = m(a)$. We feed the spectrogram features to the whisper encoder $w$ to get the audio embeddings $h^a=\{h^a_1 \ldots h^a_i \ldots h^a_n\} = w(e^a)$, where $n$ denotes the number of audio frames. We get the final audio representation $\bar{h^a}$ by temporally averaging~\cite{parthasarathy2018convolutional, pepino2021emotion, venkataramanan2019emotion, mirsamadi2017automatic} the output of the whisper encoder: $\bar{h^a}=\frac{1}{n}\sum_{i=1}^{n}h^a_i$. We encapsulate all these operations of the teacher model as $t$ henceforth so that $\bar{h^a} = t(a)$.

\begin{figure*}[!tbh]
\centering
\includegraphics[width=\textwidth]{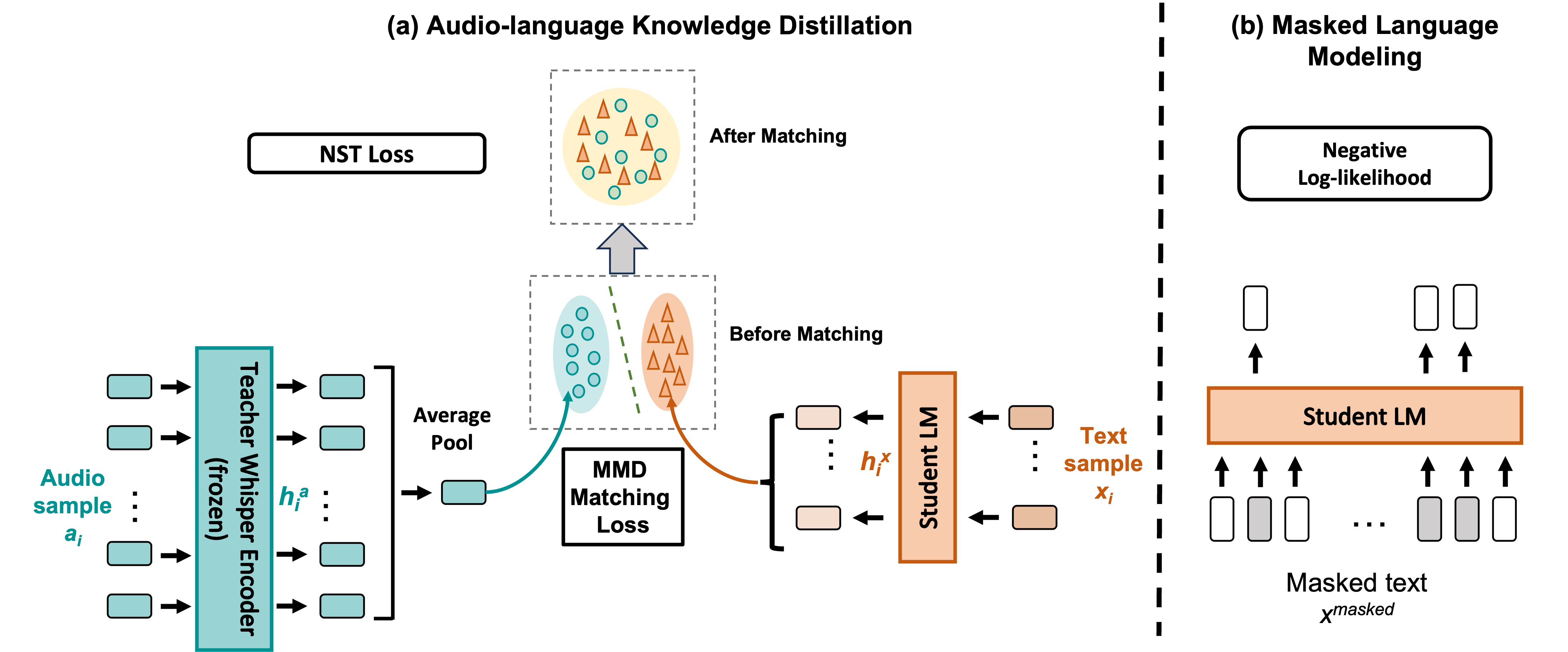}
\caption{Overview of \studentNST\  model: (a) Neuron Selectivity Transfer (NST) knowledge distillation loss for transferring sequential neuron activation patterns from the teacher to student model by 
minimizing 
the maximum mean discrepancy (MMD) between them and (b) masked language modeling (MLM) for the student model.}
\label{fig:nst}
\end{figure*}

\subsection{Student Model} 
\label{sec:student}
Our student model is constructed with the same architecture as the 12 layers of BERT~\citep{devlin2018bert} model. For each audio sample $a$, we have a transcribed text $x$; we tokenize $x$ and prepend a special token: \texttt{[CLS]}, which serves to represent the entire sentence. The student model, $s$ then takes this input, $x$, and produces a contextualized representation, $h^x = s(x) = \{h^x_{\texttt{[CLS]}}, h^x_1~\ldots ~h^x_{|x|}\}$.

\vspace{2mm}
We also include the conventional masked language modeling (MLM) objective during training
~\citep{devlin2018bert}. This objective involves replacing a certain percentage (15\%) of tokens within $x$ with a special token, \texttt{[MASK]}, resulting in a masked text, denoted as $x^{masked}$, which maintains the same length as the original sequence. Subsequently, the model utilizes $x^{masked}$ as input and learns to predict the tokens by minimizing the negative log-likelihood: 
\begin{multline}
\label{equ:mlm}
\lossMLM(x, x^{masked}) \\ = 
- \sum_{i\in Mask} \log p (x_i | x^{masked})  
\end{multline}
where $Mask$ 
is the indices of masked tokens.

The MLM loss helps the student model learn contextualized representation of sentences whereas the knowledge distillation objective ensures that this contextualized representation is aligned with the corresponding acoustic and paralinguistic features (guided by the teacher).


\subsection{Knowledge Distillation Objectives}
Here, we discuss the KD objectives used to transfer knowledge from the teacher (Whisper) model (Section~\ref{sec:teacher}) to the student model (Section~\ref{sec:student}).

For each sample in the batch, we pass the audio through the Whisper encoder to receive audio embeddings and take their average across timestamps, $\bar{h^a}$ (discussed in Section~\ref{sec:teacher}) and calculate KD losses with the text token embeddings ($h^x$).
The primary reason for taking the mean audio embedding (instead of individual audio tokens) for  calculating the KD loss  is due to the unavailability of utterance level alignment between the audio and text tokens in most training datasets. Interestingly, this is also an 
issue in vision-language knowledge distillation; ~\citet{tang2021vidlankd} also took the same approach to generate a mean video token embedding to optimize their KD objectives. The text samples can be of variable length (between 3-108 tokens in our training dataset); we only calculate the KD losses between the mean audio representation and valid token embeddings (padding tokens are ignored). Moreover, the MLM loss is computed exclusively on the masked positions, while the KD losses are calculated by considering all valid hidden states. 
During the 
knowledge distillation process, we keep the weights of the teacher model frozen for faster training and lower memory footprint.

Our method can use either of two 
KD objectives: 
Neuron Selectivity Transfer (NST)~\citep{huang2017like} and 
Contrastive Representation Distillation (CRD)~\citep{tian2019contrastive}. 

\begin{figure*}[!tbh]
\centering
\includegraphics[width=\textwidth]{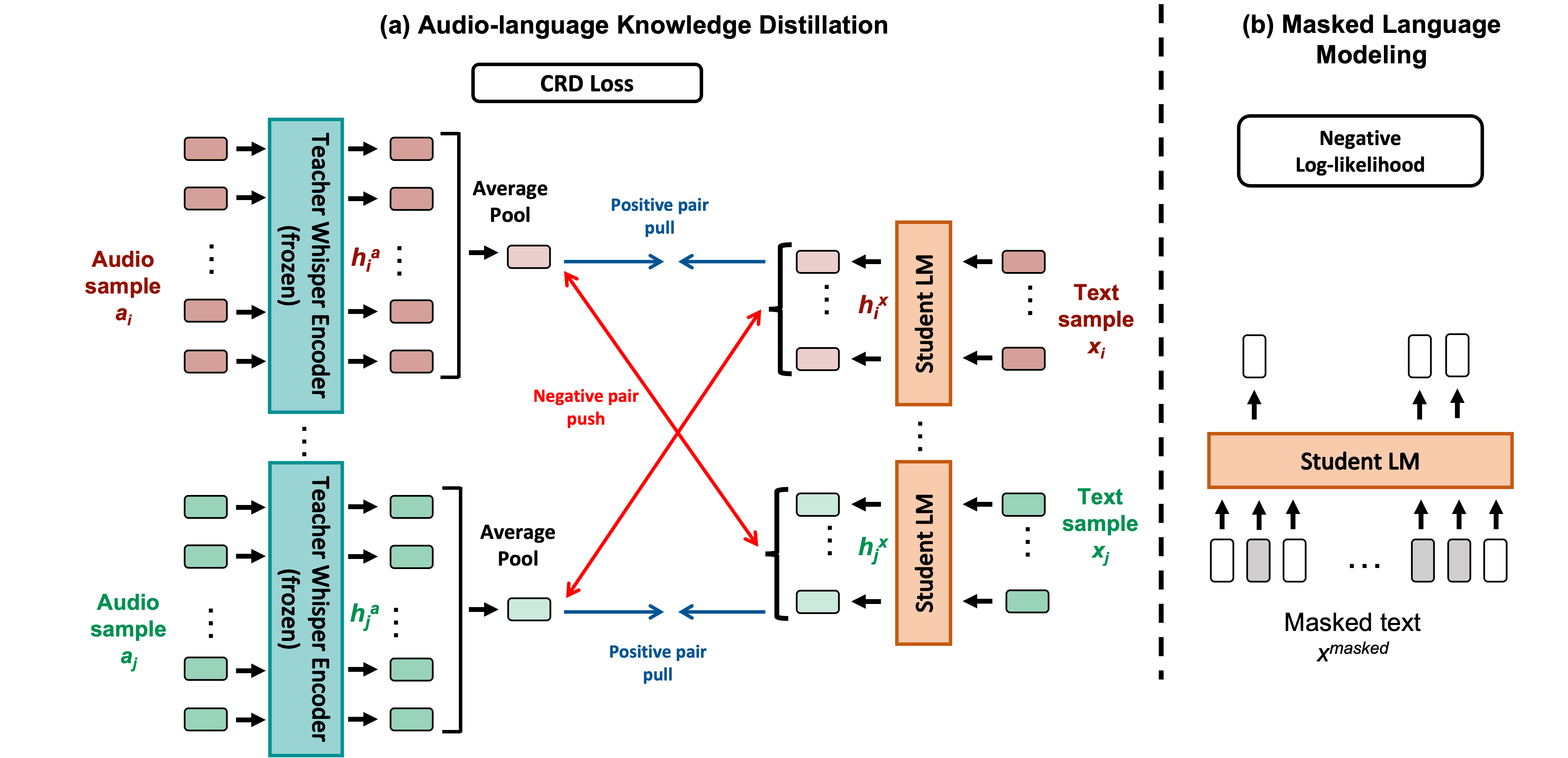}
\caption{Overview of \studentCRD\ model. (a) Contrastive Representation Distillation (CRD) knowledge distillation loss for maximizing mutual information between teacher and student representations with contrastive learning using positive and negative pairs within a batch (b) masked language modeling (MLM) for the student model.}
\label{fig:crd}
\end{figure*}

\paragraph{Neuron Selectivity Transfer (NST)}
was proposed to align the distribution of neuron selectivity (activation pattern with respect to each inputs) between student and teacher within a single modality (computer vision). 
\citet{huang2017like} hypothesized that 
each neuron in a deep neural network tends to capture specific patterns. If a specific neuron is activated for a certain set of inputs, it can lead to a clustering behavior. Such behavior provides an explanation for the final prediction of the teacher model and hence the authors proposed to teach/transfer these heatmaps-like spatial activation patterns to the student neurons from the teacher's.
\vspace{1ex}\\
\indent To adapt NST to our case, we transfer the sequential activation patterns of $t(a) \in \mathbb{R}^{d}$ to $s(x) \in \mathbb{R}^{|x|\times d}$. Given input audio $a$ and text $x$, and hidden state dimension $d$, $t(a)$ is the average of the final hidden states of the teacher-whisper model and $s(x)$ is the final hidden state (with $n$ tokens) of the student language model. Following~\citet{huang2017like} we use squared maximum mean discrepancy (MMD)~\citep{gretton2012kernel} which uses the kernel trick to measure and minimize the distance between the activation patterns of student neurons $\{s(x)_{*, i}\}^d_{i=1}$ and the teacher neurons $\{t(a)_{*, j}\}^d_{j=1}$:

\begin{equation}
\begin{split}
\label{equ:mmd}
\lossNST (a, x) &= \lossMMD^2(a, x) \\ 
&= \frac{1}{d^2} \sum_{i=1}^{d} \sum_{i'=1}^{d} k [s(x)_{*, i}; s(x)_{*, i'} ] \\
&+ \frac{1}{d^2} \sum_{j=1}^{d} \sum_{j'=1}^{d} k [t(a)_{*, j}; t(a)_{*, j'} ] \\
&- \frac{2}{d^2} \sum_{i=1}^{d} \sum_{j=1}^{d} k [s(x)_{*, i}, t(a)_{*, j} ]
\end{split}
\end{equation}

We chose to use a polynomial kernel $k[s; t] = (s^\top t + c)^d$ with $d = 2$ and $c = 0$ due to its superior performance over other kernels in~\citep{huang2017like}'s experiments. 
Since $t(a) \in \mathbb{R}^{d}$ is the average of the audio embeddings and $s(x) \in \mathbb{R}^{|x|\times d}$, we use an average of the MMD distances over all valid text tokens in the batch when computing the loss per minibatch. 
Figure \ref{fig:nst} shows the \studentNST\  model in our system.


\paragraph{Contrastive Representation Distillation (CRD)}
was specifically designed for multi-modal knowledge distillation. 
\citet{tian2019contrastive} first proposed CRD to maximize the mutual information between the teacher and student representations with contrastive learning. The authors conjecture that the contrastive objective better transfers all the information in the teacher's representation, rather than only assuming the teachers embedding dimensions are conditionally independent and only trying to transfer those representations.
Just like before, given input audio $a$ and text $x$, $t(a) \in \mathbb{R}^{d}$ and $s(x) \in \mathbb{R}^{|x|\times d}$ are teacher and student representations respectively.
From the joint distribution of teacher and student representations, we get one positive pair for every $N$ (batch-size) negative pairs that are drawn from the product of the marginal distributions. 
To maximize the lower bound of the mutual information between $s(x)$ and $t(a)$, we have to minimize the following: 
\begin{multline}
\label{equ:crd}
    \lossCRD = - \mathbb{E}_{(\mathbf{s},\mathbf{t}|positive)}[\log{\phi}(s(x), t(a))] 
    \\ - N \cdot \mathbb{E}_{(\mathbf{s},\mathbf{t}|negative)}[\log{(1-\phi}(s(x), t(a))]
\end{multline}
where
\begin{equation}
\label{equ:infonce}
    \phi(s(x), t(a)) = \frac{\exp^{s(x)^\top~t(a)/\tau}}
    {\exp^{s(x)^\top~t(a)/\tau} + \frac{N}{M}}
\end{equation} 
Here, $M$ is the cardinality of the dataset and $\tau$ is a temperature that adjusts the concentration level. \citet{tian2019contrastive} suggest that a large $N$ leads to a tighter lower bound for the mutual information, hence we opt for a large batch size (>=256) in our implementation.
Just like with the $\lossNST$, for each minibatch, we average this loss over the valid $x$ tokens.   
Figure \ref{fig:crd} shows the \studentCRD\  model.

\section{Experimental Setup}
\subsection{Datasets and Evaluation Metrics}

\paragraph{Dataset for Audio-Language Knowledge Distillation}
For the cross-modal knowledge distillation, we utilize the People's Speech~\citep{galvez2021people} dataset. This dataset comprises conversational English speech recordings sourced from the internet, ensuring the usage of appropriately licensed audio data with corresponding transcriptions. While the complete dataset encompasses \texttt{30,000+} hours of speech, it should be noted that due to licensing constraints, the available portion of the dataset suitable for academic and commercial purposes is limited to \texttt{8,273} hours. The \texttt{`clean'} portion of the training data consists of \texttt{5,895} hours of audio and this is the dataset we use for KD.
The audio recordings are divided into \texttt{15}-second chunks, resulting in \texttt{1,501,271} audio samples. Each audio sample is paired with its corresponding transcribed text (between \texttt{3-108} tokens per sample), which consists of \texttt{65,977,589} tokens in total. It is important to note that the total number of tokens in this dataset is considerably smaller compared to the token count of traditional text corpora used for training models like BERT. 
More specifically, this dataset only has 1.65\% of the total number of tokens compared to a combined Wikipedia and book corpus dataset (approx. 4B tokens) that is used to train BERT. However, this is still the largest dataset publicly available for spoken audio clips and transcription. 

\paragraph{Dataset for Downstream Fine-tuning \& Evaluation Tasks}
We conducted the evaluation of our model on the CMU-MOSEI dataset \citep{zadeh2018multimodal}, employing sentiment analysis and emotion recognition tasks. The CMU-MOSEI dataset comprises \texttt{23,454} video+audio clips of movie reviews sourced from YouTube. For our evaluation, we only utilized the transcribed text extracted from these clips. We specifically chose this dataset due to its public accessibility and the fact that it provides emotion/sentiment annotations, while also possessing a similar token length (\texttt{<128}) per sample compared to our pre-training dataset.

\paragraph{Evaluation Metrics}

In the CMU-MOSEI dataset, human annotators have assigned sentiment scores ranging from -3 (strongly negative) to 3 (strongly positive) for each sample. To assess the performance of our model, we utilize a range of metrics that align with those employed in prior studies~\citep{zadeh2018multimodal}. 
These metrics include the 7-class accuracy ($\texttt{Sentiment}^\texttt{7}$) for sentiment score classification within the -3 to 3 range, 
the 5-class accuracy ($\texttt{Sentiment}^\texttt{5}$), 
the 3-class accuracy (\texttt{positive/negative/neutral}) ($\texttt{Sentiment}^\texttt{3}$), 
the binary accuracy (\texttt{positive/negative} with or without \texttt{neutral}) 
($\texttt{Sentiment}^\texttt{2}_{\texttt{w\_neutral}}$ and $\texttt{Sentiment}^\texttt{2}_{\texttt{w/o\_neutral}}$), 
the mean absolute error (\texttt{MAE}) of the sentiment score, and the Pearson's correlation coefficient ($\rho$) between the model's predictions and the human assessments. Additionally, the dataset includes annotations for six emotions for binary classification: {\textit{happiness}, \textit{sadness}, \textit{surprise}, \textit{anger}, \textit{fear}, \textit{disgust}}. We also report accuracy measures for the emotion recognition tasks.

\begin{table*}[!thb]
  \centering
  \resizebox{\textwidth}{!}{%
  \begin{tabular}{@{}lcccccc@{}}
  \toprule \toprule
    & \texttt{Sentiment\textsuperscript{7}} & \texttt{Sentiment\textsuperscript{5}} & \texttt{Sentiment\textsuperscript{3}} & $\texttt{Sentiment}^\texttt{2}_{\texttt{w/o\_neutral}}$ & $\texttt{Sentiment}^\texttt{2}_{\texttt{w\_neutral}}$ & Average \\
   Model (\# tokens) & $\texttt{Accuracy}\pm~\sigma$ & $\texttt{Accuracy}\pm~\sigma$ & $\texttt{Accuracy}\pm~\sigma$ & $\texttt{Accuracy}\pm~\sigma$ & $\texttt{Accuracy}\pm~\sigma$ & \texttt{Accuracy}\\ 
  \midrule \midrule
  \BERT (4B)   & $47.80 \pm 0.29$ & $51.89 \pm 0.19$ & $69.45 \pm 0.23$ & $85.88 \pm 0.40$ & $76.78 \pm 0.19$ & $66.36$\\
  \wikiBERT (2.9B) & $46.03 \pm 0.60$ & $49.89 \pm 0.45$ & $67.25 \pm 0.25$ & $83.81 \pm 0.37$ & $75.03 \pm 0.29$ & $64.40$\\
  \spokenBERT (66M) & $45.57 \pm 0.16$ & $49.60 \pm 0.33$ & $66.91 \pm 0.16$ & $83.65 \pm 0.34$ & $74.26 \pm 0.25$ & $64.00$\\ \midrule
  \texttt{Student\textsubscript{NST}} (66M)  & $46.73 \pm 0.40$ & $51.44 \pm 0.01$ & $68.83 \pm 0.26$ & $84.94 \pm 0.17$ & $76.52 \pm 0.09$ & $65.69$ \\
  \texttt{Student\textsubscript{CRD}} (66M)  & $47.09 \pm 0.01$ & $51.44 \pm 0.20$ & $68.36 \pm 0.10$ & $85.16 \pm 0.11$ & $76.28 \pm 0.26$ & $65.67$ \\ 
  \bottomrule \bottomrule
  \end{tabular}%
  }
  \caption{Average accuracy (in \%) comparison between the models in sentiment label prediction (classification) tasks. Here, $\sigma$ denotes the standard deviation over 5 trials.}
  \label{tab:sentiment_label}
  \end{table*}

\paragraph{Baselines}

We compared our model's performance against three baselines: (i) \BERT~\citep{devlin2018bert}, (ii) \wikiBERT~\citep{tang2021vidlankd, tan2020vokenization}, and (iii) \spokenBERT, all having identical transformer architecture (12 layers) and training hyper-parameters,  with our proposed model. \BERT~is trained on a combination of the English Wikipedia and Book corpus, totaling approximately 4 billion tokens, while \wikiBERT~is trained solely on the English Wikipedia, which consists of around 2.9 billion tokens. On the other hand, \spokenBERT~is trained on the transcription of the People's Speech dataset~\citep{galvez2021people}, containing 66 million tokens. We selected these three baselines to showcase the impact of token quantity (millions vs billions) and characteristics (such as spoken colloquial language vs. written formal language), as well as to highlight the effects of incorporating audio information into spoken language text corpora.

\subsection{Implementation Details}
During the pre-processing step, we prepare our \texttt{15s}, \texttt{16kHz} audio samples for the Whisper model by padding them to match the expected \texttt{30s} audio sequence length. Subsequently, the padded audio samples are transformed into FFT spectrograms, consisting of a sequence of \texttt{25ms} frames with a \texttt{10ms} hop length and \texttt{400} FFT features. From these spectrograms, we extract \texttt{80} log-Mel features using the  \texttt{slaney} scale, resulting in audio features of shape \texttt{80x3000}. The transformed Mel-frames are then passed through the Whisper encoder, producing audio embeddings of shape \texttt{768x1500}. To remove the corresponding padding introduced earlier, we discard the last \texttt{750} frames from the audio embeddings. Finally, we obtain the final audio embedding by performing average pooling. 


\begin{table}[!t]
\resizebox{0.48\textwidth}{!}{%
\begin{tabular}{@{}lcc@{}}
\toprule \toprule
& \multicolumn{2}{c}{Sentiment\textsuperscript{score}} \\ 
Model (\# tokens) & \multicolumn{1}{c}{$\downarrow\texttt{MAE} \pm \sigma$} & \multicolumn{1}{c}{$\uparrow\rho \pm \sigma$ } \\ \midrule \midrule
\BERT (4B)      & $0.5453 \pm 0.0015$   & $76.99 \pm 0.09$ \\
\wikiBERT (2.9B)    & $0.5858 \pm 0.0029$   & $72.87 \pm 0.24$ \\
\spokenBERT (66M) &
 $0.5903 \pm 0.0019$ &
 $72.00 \pm 0.18$ \\ \midrule
\texttt{Student\textsubscript{NST}} (66M) & $0.5590 \pm 0.0012$  & $74.61 \pm 0.19$ \\
\texttt{Student\textsubscript{CRD}} (66M) & $0.5614 \pm 0.0003$  & $74.74 \pm 0.12$ \\ \bottomrule\bottomrule
\end{tabular}%
}
\caption{Average mean absolute error (MAE) and Pearson's correlation coefficient ($\rho$, in \%) for the models in sentiment score prediction (regression) tasks. Here, $\sigma$ denotes the standard deviation over 5 trials; $\downarrow$ and $\uparrow$ denote lower and higher values are preferred, respectively.}
\label{tab:sentiment_score}
\end{table}

To improve computational efficiency, we modified the \texttt{BertTokenizer} and \texttt{BertModel} to process text samples up to 128 tokens. 
We trained the student models using  the \texttt{AdamW} optimizer with a learning rate of \texttt{1e-4}, \texttt{10k} warm-up steps, \texttt{40} training epochs, a batch size of \texttt{256}, and a weight decay of \texttt{0.01} in a mixed precision training (\texttt{bf16}) regimen. 
For downstream tasks we report results on the test sets. 
We fine-tuned on these tasks for \texttt{3} epochs, with a learning rate of \texttt{2e-5} and a batch size of \texttt{32}.
For the $\lossCRD$, we set the temperature parameter $\tau$ to \texttt{0.01} (cf. Equation~\ref{equ:infonce}). For both $\lossCRD$ and $\lossNST$, we set the parameter $\gamma$ to \texttt{1.0} (cf. Equation~\ref{eqn:total_loss}).
We implemented our models using \texttt{PyTorch} and \texttt{HuggingFace} libraries, and the training process was conducted on Nvidia GeForce RTX 4090 GPUs. Training a single run of \texttt{Student\textsubscript{NST}} and \texttt{Student\textsubscript{CRD}} models on a single GPU took approximately 7 and 9 days, respectively.
During the evaluation, we conducted each experiment with 5 different seeds and reported the average results and the standard deviations, $\sigma$.

\section{Results and Analysis}

\begin{table*}[!htb]
\resizebox{\textwidth}{!}{%
\begin{tabular}{lccccccc}
\toprule \toprule
 & Happiness & Sadness & Surprise & Anger & Fear & Disgust & Average\\
Model (\# tokens) & $\texttt{Accuracy}\pm~\sigma$ & $\texttt{Accuracy}\pm~\sigma$ & $\texttt{Accuracy}\pm~\sigma$ & $\texttt{Accuracy}\pm~\sigma$ & $\texttt{Accuracy}\pm~\sigma$ & $\texttt{Accuracy}\pm~\sigma$ &
$\texttt{Accuracy}$ \\\midrule \midrule
\BERT (4B) &
  $67.93 \pm 0.23$ &
  $\mathbf{75.18 \pm 0.26}$ &
  $90.45 \pm 0.11$ &
  $78.33 \pm 0.20$ &
  $91.71 \pm 0.04$ &
  $84.91 \pm 0.23$ &
  $81.42$ \\
\wikiBERT (2.9B) &
  $67.02 \pm 0.81$ &
  $73.69 \pm 0.41$ &
  $89.50 \pm 0.34$ &
  $78.19 \pm 0.06$ &
  $91.08 \pm 0.06$ &
  $84.68 \pm 0.25$ &
  $80.69$ \\
\spokenBERT (66M) &
  $66.55 \pm 0.22$ &
  $74.16 \pm 0.02$ &
  $89.80 \pm 0.10$ &
  $77.76 \pm 0.25$ &
  $91.43 \pm 0.17$ &
  $84.24 \pm 0.16$ &
  $80.66$ \\ \midrule \midrule
\texttt{Student\textsubscript{NST}} (66M) &
  $\mathbf{68.22 \pm 0.21}$ &
  $74.70 \pm 0.31$ &
  $90.83 \pm 0.24$ &
  $\mathbf{78.48 \pm 0.20}$ &
  $\mathbf{92.74 \pm 0.02}$ &
  $\mathbf{85.36 \pm 0.04}$ &
  $\mathbf{81.72}$ \\
\texttt{Student\textsubscript{CRD}} (66M) &
  $67.87 \pm 0.01$ &
  $74.76 \pm 0.22$ &
  $\mathbf{91.49 \pm 0.42}$ &
  $78.28 \pm 0.10$ &
  $92.40 \pm 0.21$ &
  $85.24 \pm 0.37$ &
  81.67\\ \bottomrule \bottomrule
\end{tabular}%
}
\caption{Average accuracy (in \%) comparison between the models in emotion recognition (binary classification) tasks. Here, $\sigma$ denotes the standard deviation over 5 trials.}
\label{tab:emotion}
\end{table*}

As there is a large difference in the number of tokens used to train our model (66M) and the BERT baseline (4B), as well as the characteristics of the datasets (formal written text vs. informal spoken text), we first investigate whether the People's Speech dataset we used contains enough tokens and desired characteristics to train a 12 layer transformer model from scratch for emotion and sentiment prediction tasks. The second part of our analysis studies whether leveraging the audio features during pretraining improves the downstream tasks.

\subsection{Impact of pretraining dataset}

We first observe a trend in Tables \ref{tab:sentiment_label} and \ref{tab:sentiment_score}, where a decrease in the number of tokens from 4B to 66M corresponds to a decline in performance across the three baselines. Specifically, transitioning from \BERT~to \wikiBERT~(a 27.5\% reduction in tokens from 4B), we observe a decrease in average sentiment classification accuracy by 1.96\% (Table \ref{tab:sentiment_label}), an increase in sentiment regression MAE by 0.04, and a 4.12\% decrease in $\rho$ (Table \ref{tab:sentiment_score}). Despite a substantial reduction in the total number of tokens (97.72\%) from \wikiBERT~to \spokenBERT, the decline in performance is relatively modest (0.4\% in accuracy). 
It is intriguing to consider how a model trained on 44 times fewer tokens can perform competitively against a substantially trained model. Our hypothesis is that this phenomenon may be attributed to the characteristics of the People's Speech dataset, which encompasses a rich and diverse vocabulary for informal spoken language. This aligns with the characteristics of CMU-MOSEI dataset which is also based on spoken language. 
We observe a similar trend in emotion classification (Table \ref{tab:emotion}), where the transition from \BERT~to \wikiBERT~results in an average drop of 0.73\% in classification accuracy. However, the shift from \wikiBERT~to \spokenBERT~yields a mere 0.03\% decline in the same metric, despite the substantial reduction in training tokens. This observation not only justifies the usage of the People's Speech dataset for training the language model but also motivates the exploration of integrating audio features to further improve the sentiment prediction and emotion classification performance.

\subsection{Impact of audio information}

Upon integrating audio features through our knowledge distillation approach, we observe a significant improvement over \spokenBERT, the text-only model trained on the same token set as our \texttt{Student} models, across all evaluation tasks. Specifically, in sentiment classification (Table \ref{tab:sentiment_label}), \studentNST~and \studentCRD~achieve an average improvement of 1.69\% and 1.67\%, compared to \spokenBERT\  respectively. In the sentiment score prediction task (Table \ref{tab:sentiment_score}), the MAE decreases by 0.0314 and 0.0290, while $\rho$ increases by 2.62\% and 2.74\%, respectively. This positive trend extends to emotion classification, where \studentNST~and \studentCRD~exhibit an average improvement of 1.06\% and 1.02\%, respectively. These findings affirm that the inclusion of additional audio information strengthens the language model for understanding spoken transcripts. 

Upon closer examination, we note that the performance gains achieved by \studentNST~and \studentCRD~ surpass those of \wikiBERT~across all tasks. Specifically, in average sentiment classification (Table \ref{tab:sentiment_label}), these models demonstrate accuracy improvements of 1.29\% and 1.27\% over \wikiBERT. In the sentiment score regression task (Table \ref{tab:sentiment_score}), the corresponding MAE decreases are 0.0268 and 0.0244, while $\rho$ increases by 1.74\% and 1.87\%. Furthermore, in emotion classification tasks (Table \ref{tab:emotion}), we observe average improvements of 1.03\% and 0.98\% for \studentNST~and \studentCRD, respectively. These results are particularly encouraging, as the model trained on 66M tokens demonstrates an enhancement in overall performance metrics compared to the model trained on 2.9B tokens.
When conducting a performance comparison with \BERT, we observe that the student models begin to lag behind in sentiment prediction tasks. However, in emotion prediction tasks, the student models achieve an average improvement of up to 0.3\% (\studentNST).

In summary, both our student models, \studentNST~and \studentCRD, exhibit superior performance compared to the text-based model \spokenBERT, as well as \wikiBERT\ (which is trained with 44 times the number of text tokens). Moreover, in emotion classification, our student models even outperform the fully trained \BERT~model. The low standard deviation across all trials offers compelling evidence of the statistical significance of these findings.
It is worth noting that $\lossCRD$ and $\lossNST$ demonstrate similar performance without a clear winner.

\section{Conclusion}
We developed a method to train a large language model for analyzing spoken transcripts by leveraging audio and paralinguistic features, without requiring audio at inference time. We developed a knowledge distillation approach that leverages OpenAI Whisper's pre-trained speech embeddings as a teacher model and transfers its knowledge to our student models.  By including audio information, our models can outperform traditional language models, even with a small fraction of the tokens.

\section*{Limitations}
We did not tune the $\gamma$ (relative strength between MLM and KD loss) and $\tau$ (temperature in CRD loss) hyper-parameters due to our limited computation budget.
For the same reason, we were limited to performing experiments on 12 layers of BERT instead of BERT-large 24 layers or Whisper large.
It would be ideal to train BERT-large on the full 30K hours of the People Speaking dataset, while we were only able to train on 5895 hours of data. 
Moreover, while it would be preferable to validate our results on more downstream tasks that have similar token lengths (<128) to our pre-training dataset,  CMU-MOSEI was the only such dataset available to us. Alternatively, pre-training with longer audio sequences (e.g., 30s vs. our 15s) and corresponding text tokens (>128) would enable us to explore more downstream tasks, e.g., UR-FUNNY~\citep{hasan2019ur}, sarcasm detection~\citep{castro2019towards} task.
It would also be worth investigating the models with additional KD objectives, and an extension of our method that uses a combination of KD objectives. 

\bibliography{acl,custom}
\bibliographystyle{acl_natbib}
\end{document}